\documentclass[11pt,a4paper]{article}
\usepackage{times,latexsym}
\usepackage{url}
\usepackage[T1]{fontenc}

\usepackage[acceptedWithA]{tacl2018v2}
\usepackage[]{tacl2018v2}
\usepackage{graphicx}
\usepackage{tabularx}
\usepackage{booktabs}
\usepackage{multirow}
\usepackage{url}

\usepackage{xspace,mfirstuc,tabulary}

\newif\iftaclinstructions
\taclinstructionsfalse % AUTHORS: do NOT set this to true
\iftaclinstructions
\renewcommand{\confidential}{}
\renewcommand{\anonsubtext}{(No author info supplied here, for consistency with
TACL-submission anonymization requirements)}
\newcommand{\instr}
\fi

\iftaclpubformat % this "if" is set by the choice of options

\else

\fi

\newcommand{\BB}[1]{BERT\textsubscript{BASE}}
\newcommand{\BL}[1]{BERT\textsubscript{LARGE}}
\newcommand{\fk}[1]{CPRAG-102}
\newcommand{\rr}[1]{ROLE-88}
\newcommand{\nkf}[1]{NEG-136}
\newcommand{\fs}[1]{NEG-136-SIMP}
\newcommand{\nk}[1]{NEG-136-NAT}

\title{What BERT is not: Lessons from a new suite of psycholinguistic diagnostics for language models}

\author{
Allyson Ettinger \\
Department of Linguistics \\
University of Chicago\\
  {\sf aettinger@uchicago.edu} \\
}

\date{}

\begin{document}
\maketitle
\begin{abstract}
Pre-training by language modeling has become a popular and successful approach to NLP tasks, but we have yet to understand exactly what linguistic capacities these pre-training processes confer upon models. In this paper we introduce a suite of diagnostics drawn from human language experiments, which allow us to ask targeted questions about information used by language models for generating predictions in context. As a case study, we apply these diagnostics to the popular BERT model, finding that it can generally distinguish good from bad completions involving shared category or role reversal, albeit with less sensitivity than humans, and it robustly retrieves noun hypernyms, but it struggles with challenging inference and role-based event prediction---and in particular, it shows clear insensitivity to the contextual impacts of negation.
%finding that while this model is able to use some linguistic distinctions effectively for guiding predictions, the model struggles with challenging inferences and event predictions, shows generally weaker sensitivity to linguistic distinctions than do humans, and in particular shows a stark inability to adjust to the contextual impacts of negation.
\end{abstract}

\section{Introduction}

Pre-training of NLP models with a language modeling objective has recently gained popularity as a precursor to task-specific fine-tuning. Pre-trained models like BERT~\cite{devlin2018bert} and ELMo~\cite{peters2018deep} have advanced the state of the art in a wide variety of tasks, suggesting that these models acquire valuable, generalizable linguistic competence during the pre-training process.  However, though we have established the benefits of language model pre-training, we have yet to understand what exactly about language these models learn during that process. 

This paper aims to improve our understanding of what language models (LMs) know about language, by introducing a set of diagnostics targeting a range of linguistic capacities, drawn from human psycholinguistic experiments. Because of their origin in psycholinguistics, these diagnostics have two distinct advantages: they are carefully controlled to ask targeted questions about linguistic capabilities, and they are designed to ask these questions by examining word predictions in context, which allows us to study LMs without any need for task-specific fine-tuning.

Beyond these advantages, our diagnostics distinguish themselves from existing tests for LMs in two primary ways. First, these tests have been chosen specifically for their capacity to reveal insensitivities in predictive models, as evidenced by patterns that they elicit in human brain responses. 
%These particular test sets are selected because in human experiments they demonstrate an ability to separate more and less sophisticated predictive mechanisms, presenting pitfalls for mechanisms that use heuristics. 
Second, each of these tests targets a set of linguistic capacities that extend beyond the primarily syntactic focus seen in existing LM diagnostics---we have tests targeting commonsense/pragmatic inference, semantic roles and event knowledge, category membership, and negation. Each of our diagnostics is set up to support tests of both word prediction accuracy and sensitivity to distinctions between good and bad context completions. Although we focus on the BERT model here as an illustrative case study, these diagnostics are applicable for testing of any language model.

This paper makes two main contributions. First, we introduce a new set of targeted diagnostics for assessing linguistic capacities in language models.\footnote{All test sets and experiment code are made available here: \url{https://github.com/aetting/lm-diagnostics}}\label{fn:link} Second, we apply these tests to shed light on strengths and weaknesses of the popular BERT model. We find that BERT struggles with challenging commonsense/pragmatic inferences and role-based event prediction, that it is generally robust on within-category distinctions and role reversals, but with lower sensitivity than humans, and that it is very strong at associating nouns with hypernyms. Most strikingly, however, we find that BERT fails completely to show generalizable understanding of negation, raising questions about the aptitude of LMs to learn this type of meaning.

\section{Motivation for use of psycholinguistic tests on language models}

It is important to be clear that in using these diagnostics, we are not testing whether LMs are psycholinguistically plausible. 
We are using these tests simply to examine LMs' general linguistic knowledge, specifically by asking what information the models are able to use when assigning probabilities to words in context. These psycholinguistic tests are well-suited to asking this type of question because a) the tests are designed for drawing conclusions based on predictions in context, allowing us to test LMs in their most natural setting, and b) the tests are designed in a controlled manner, such that accurate word predictions in context depend on particular types of information. In this way, these tests provide us with a natural means of diagnosing what kinds of information LMs have picked up on during training.

Clarifying the linguistic knowledge acquired during LM-based training is increasingly relevant as state-of-the-art NLP models shift to be predominantly based on pre-training processes involving word prediction in context. In order to understand the fundamental strengths and limitations of these models---and in particular, to understand what allows them to generalize to many different tasks---we need to understand what linguistic competence and general knowledge this LM-based pre-training makes available (and what it does not). The importance of understanding LM-based pre-training is also the motivation for examining pre-trained BERT, as we do in the present paper, despite the fact that the pre-trained form is typically used only as a starting point for fine-tuning. Because it is the pre-training that seemingly underlies the generalization power of the BERT model, allowing for simple fine-tuning to perform so impressively, it is the pre-trained model that presents the most important questions about the nature of generalizable linguistic knowledge in BERT. 
%This is how we can gain a better understanding of what pre-training contributes to make BERT so effective (and, importantly for our purposes, where pre-training is limited).

\section{Related Work}

This paper contributes to a growing effort to better understand the specific linguistic capacities achieved by neural NLP models. Some approaches use fine-grained classification tasks to probe information in sentence embeddings~\cite{adi2016fine,conneau2018you,ettinger2018assessing}, or token-level and other sub-sentence level information in contextual embeddings~\cite{tenney2018you,peters2018dissecting}. 
Some of this work has targeted specific linguistic phenomena such as function words~\cite{kim2019probing}. 
Much work has attempted to evaluate systems' overall level of ``understanding'', often with tasks such as semantic similarity and entailment~\cite{wang2018glue,bowman2015large,agirre2012semeval,dagan2005pascal,bentivogli2016sick}, and additional work has been done to design curated versions of these tasks to test for specific linguistic capabilities~\cite{dasgupta2018evaluating,poliak2018collecting,mccoy2019right}.  Our diagnostics complement this previous work in allowing for direct testing of language models in their natural setting---via controlled tests of word prediction in context---without requiring probing of extracted representations or task-specific fine-tuning. 

More directly related is existing work on analyzing linguistic capacities of language models specifically. This work is particularly dominated by testing of syntactic awareness in LMs, and often mirrors the present work in employing targeted evaluations modeled after psycholinguistic tests~\cite{linzen2016assessing,gulordava2018colorless,marvin2018targeted,wilcox2018rnn,chowdhury2018rnn,futrell2019neural}. These analyses, like ours, typically draw conclusions based on LMs' output probabilities. Additional work has examined the internal dynamics underlying LMs' capturing of syntactic information, including testing of syntactic sensitivity in different components of the LM and at different timesteps within the sentence \cite{giulianelli2018under}, or in individual units \cite{lakretz2019emergence}.

%examining internal dynamics of the LM in tracking syntactic phenomenon of number agreement, examining contributions of individual units \cite{lakretz2019emergence}.

The above work analyzing language models focuses heavily on syntactic competence---semantic phenomena like negative polarity items are tested in some studies~\cite{marvin2018targeted,jumeletlanguage}, but the tested capabilities in these cases are still firmly rooted in the notion of detecting structural dependencies. In the present work we expand beyond the syntactic focus of the previous literature, testing for capacities including commonsense/pragmatic reasoning, semantic role and event knowledge, category membership, and negation---while continuing to use controlled, targeted diagnostics. Our tests are also distinct in eliciting a very specific response profile in humans, creating unique predictive challenges for models, as described below. 

We further deviate from previous work analyzing LMs in that we not only compare word probabilities---we also examine word prediction accuracies directly, for a richer picture of models' specific strengths and weaknesses. Some previous work has used word prediction accuracy as a test of LMs' language understanding---the LAMBADA dataset~\cite{paperno2016lambada}, in particular, tests models' ability to predict the final word of a passage, in cases where the final sentence alone is insufficient for prediction. However, while LAMBADA presents a challenging prediction task, it is not well-suited to ask targeted questions about types of information used by LMs for prediction---unlike our tests, LAMBADA is not controlled to isolate and test the use of specific types of information in prediction. Our tests are thus unique in taking advantage of the additional information provided by testing word prediction accuracy, while also leveraging the benefits of controlled sentences that allow for asking targeted questions.

Finally, our testing of BERT relates to a growing literature examining linguistic characteristics of the BERT model itself, to better understand what underlies the model's impressive performance. \newcite{clark2019does} analyze the dynamics of BERT's self-attention mechanism, probing attention heads for syntactic sensitivity and finding that individual heads specialize strongly for syntactic and coreference relations. \newcite{lin2019open} also examine syntactic awareness in BERT by syntactic probing at different layers, and by examination of syntactic sensitivity in the self-attention mechanism. \newcite{tenney2019bert} test a variety of linguistic tasks at different layers of the BERT model. Most similarly to our work here,~\newcite{goldberg2019assessing} tests BERT on several of the targeted syntactic evaluations described above for LMs, finding BERT to exhibit very strong performance on these measures. Our work complements these approaches in testing BERT's linguistic capacities directly via the word prediction mechanism, and in expanding beyond the syntactic tests used to examine BERT's predictions in~\newcite{goldberg2019assessing}.

\begin{table*}[t]
\begin{center}
\begin{tabularx}{.95\textwidth}{X | l | l l}
\toprule
Context & Expected & Inappropriate  \\
\midrule
\emph{He complained that after she kissed him, he couldn't get the red color off his face.	He finally just asked her to stop wearing that} \_\_\_\_ & \emph{lipstick} & \emph{mascara} $|$ \emph{bracelet} \\
\emph{He caught the pass and scored another touchdown. There was nothing he enjoyed more than a good game of} \_\_\_\_ & \emph{football} & \emph{baseball} $|$ \emph{monopoly}\\
\bottomrule
\end{tabularx}
\caption{\label{tab:fksents} Example items from \fk{} dataset}
\end{center}
\vspace{-4mm}
\end{table*}

\section{Leveraging human studies}
The power in our diagnostics stems from their origin in psycholinguistic studies---the items have been carefully designed for studying specific aspects of language processing, and each test has been shown to produce informative patterns of results when tested on humans. In this section we provide relevant background on human language processing, and explain how we use this information to choose the particular tests used here.

\subsection{Background: prediction in humans}
To study language processing in humans, psycholinguists often test human responses to words in context, in order to better understand the information that our brains use to generate predictions. In particular, there are two types of predictive human responses that are relevant to us here:

%Each test has a fairly constrained set of information that models need to access and apply in order to make accurate predictions. Each test also has an insensitivity demonstrated by the N400 for a certain distinction, which we use to test the LMs' robustness to these distinctions. (Sometimes these target the same thing, sometimes not.)
%. For instance, one might present a context like the following \\
%
%\emph{The restaurant owner forgot which customer the waitress had \_\_\_\_ .} \\\\
%This is a valuable resource for studying LM-type models, since looking at predictions in context is a very natural test for models trained as LMs. In particular, tests from psycholinguistics are carefully designed to use word expectations to test specific hypotheses, which is what allows us to glean information from the test sets that we use here. 

%More specifically, we choose three particular test sets that have been shown to tease apart two different types of predictive responses in humans: the cloze response, and the N400 brain response. 

\paragraph{Cloze probability} The first measure of human expectation is a measure of the ``cloze'' response. In a cloze task, humans are given an incomplete sentence and tasked with filling their expected word in the blank. ``Cloze probability'' of a word $w$ in context $c$ refers to the proportion of people who choose $w$ to complete $c$. We will treat this as the best available gold standard for human prediction in context---humans completing the cloze task typically are not under any time pressure, so they have the opportunity to use all available information from the context to arrive at a prediction. 

\paragraph{N400 amplitude} The second measure of human expectation is a brain response known as the N400, which is detected by measuring electrical activity at the scalp (by electroencephalography). Like cloze, the N400 can be used to gauge how expected a word $w$ is in a context $c$---the amplitude of the N400 response appears to be sensitive to fit of a word in context, and has been shown to correlate with cloze in many cases~\cite{kutas1984brain}. The N400 has also been shown to be predicted by LM probabilities~\cite{frank2013word}. However, the N400 differs from cloze in being a real-time response that occurs only 400 milliseconds into the processing of a word. Accordingly, the expectations reflected in the N400 sometimes deviate from the more fully-formed expectations reflected in the untimed cloze response. 

%This deviation between cloze and N400 is a characteristic of each of the three selected test sets. This suggests that not only are these tests controlled so as to allow us to ask specific questions about the information used by LMs to generate predictions---they are also proven to have pitfalls for less sophisticated predictive mechanisms, allowing us to test models' sophistication in the face of this adversarial circumstance.
\subsection{Our diagnostic tests}

The test sets that we use here are all drawn from human studies that have revealed \emph{divergences between cloze and N400 profiles}---that is, for each of these tests, the N400 response suggests a level of insensitivity to certain information when computing expectations, causing a deviation from the fully-informed cloze predictions. We choose these as our diagnostics because they provide built-in sensitivity tests targeting the types of information that appear to have reduced effect on the N400---and because they should present particularly challenging prediction tasks, tripping up models that fail to use the full set of available information. 
%For our purposes, we might think of these as cases where the N400 is ``fooled'' by the contexts, while the humans completing the cloze task are not.\footnote{We are somewhat oversimplifying the relevant psycholinguistic results and theories. We do this because the primary purpose here is to examine the behavior of models on sets of sentences that create different challenges for prediction, and that require different capacities in order to make those predictions. We use the psycholinguistic results mostly as a guide in selecting these materials, and as inspiration for identifying what to look for in models' behavior.} We select these tests, therefore, because a) as human experiments they present targeted diagnostics, but also b) they do so in a way that teases apart more and less sophisticated predictive mechanisms, presenting unique challenges for models relying on heuristics. \td{revisit once more detail provided below}  

\section{Datasets}

Each of our diagnostics supports three types of testing: word prediction accuracy, sensitivity testing, and qualitative prediction analysis. Because these items are designed to draw conclusions about human processing, each set is carefully constructed to constrain the information relevant for making word predictions. This allows us to examine how well LMs use this target information.

For word prediction accuracy, we use the most expected items from human cloze probabilities as the gold completions.\footnote{With one exception, \nkf{}, for which we use completion truth, as in the original study.} These represent predictions that models should be able to make if they access and apply all relevant context information when generating probabilities for target words. 

For sensitivity testing, we compare model probabilities for good versus bad completions---specifically, comparisons on which the N400 showed reduced sensitivity in experiments.  This allows us to test whether LMs will show similar insensitivities on the relevant linguistic distinctions. 

Finally, because these items are constructed in such a controlled manner, qualitative analysis of models' top predictions can be highly informative about information being applied for prediction. We leverage this in our experiments below.

In all tests, the target word to be predicted falls in the final position of the provided context, which means that these tests should function similarly for either left-to-right or bidirectional LMs. Similarly, because these tests require only that a model can produce  token probabilities in context, they are equally applicable to the masked LM setting of BERT as to a standard LM. In anticipation of testing the BERT model, and to facilitate fair future comparisons with the present results, we filter out items for which the expected word is not in BERT's single-word vocabulary, to ensure that all expected words can be predicted.

%three small, targeted test sets for language models, each of which tests models' ability to use/distinguish a different type of information, and each designed for three types of informative analysis: prediction accuracy, preference of good versus distractor continuations, and analysis of preferred predictions.

It is important to acknowledge that these are small test sets, due to their origin in psycholinguistic studies. However, because these sets have been hand-designed by cognitive scientists to test predictive processing in humans, their value is in the targeted assessment that they provide with respect to information that LMs use in prediction.

We now we describe each test set in detail.

\subsection{\fk{}: commonsense and pragmatic inference}

Our first set targets commonsense and pragmatic inference, and tests sensitivity to differences within semantic category. The left column of Table~\ref{tab:fksents} shows examples of these items, each of which consists of two sentences. These items come from an influential human study by \newcite{federmeier1999rose}, which tested how brains would respond to different types of context completions, shown in the right columns of Table~\ref{tab:fksents}. 

\paragraph{Information needed for prediction} Accurate prediction on this set requires use of commonsense reasoning to infer what is being described in the first sentence, and pragmatic reasoning to determine how the second sentence relates. For instance, in Table~\ref{tab:fksents}, commonsense knowledge informs us that red color left by kisses suggests lipstick, and pragmatic reasoning allows us to infer that the thing to stop wearing is related to the complaint.
%\footnote{Unlike in the subsequent two sets, the information identified here is not the same information that was of interest to the authors in designing the materials. However, the need to use the identified information is there nonetheless, presenting an interesting high-level prediction challenge for LMs.} 
As in LAMBADA, the final sentence is generic, not supporting prediction on its own. Unlike LAMBADA, the consistent structure of these items allows us to target specific model capabilities,\footnote{To highlight this advantage, as a supplement for this test set we provide specific annotations of each item, indicating the knowledge/reasoning required to make the prediction.} and additionally, none of these items contain the target word in context,\footnote{More than 80\% of LAMBADA items contain the target word in the preceding context.} forcing models to use commonsense inference rather than coreference. Human cloze probabilities show a high level of agreement on  appropriate completions for these items---average cloze probability for expected completions is .74. 
%Reminiscent of LAMBADA, in each item the second sentence is designed to be generic, requiring incorporation of the first sentence for prediction. Unlike LAMBADA, these items are designed in a constrained way with shorter contexts and more consistent structure, allowing us to ask more targeted questions.\footnote{To emphasize this advantage, as a supplement for this this test set we provide specific annotations of each item indicating the knowledge/reasoning required to make the prediction.}

\paragraph{Sensitivity test} The~\newcite{federmeier1999rose} study found that while the inappropriate completions (e.g., \emph{mascara, bracelet}) had cloze probabilities of virtually zero (average cloze .004 and .001, respectively), the N400 showed some expectation for completions that shared a semantic category with the expected completion (e.g., \emph{mascara}, by relation to \emph{lipstick}). Our sensitivity test targets this distinction, testing whether LMs will favor inappropriate completions based on shared semantic category with expected completions. 
%\td{(This is a case in which the N400 sensitivity test targets different information from the base prediction task.)}

\paragraph{Data} The authors of the original study make available 40 of their contexts---we filter out six to accommodate BERT's single-word vocabulary,\footnote{For a couple of items, we also replace an inappropriate completion with another inappropriate completion of the same semantic category to accommodate BERT's vocabulary.} for a final set of 34 contexts, 102 total items.\footnote{Our ``item'' counts use all context/completion pairings.}

%major value: difficult prediction problems where the context is quite constraining for humans but requires synthesizing world knowledge, commonsense/pragmatic reasoning, and understanding of the basic sentence meaning, to make an accurate prediction in context. looking at the specific predictions is informative about which of those types of information the model was able to incorporate. FK furthermore comes with distractor continuations from the human N400 studies, which were shown in some conditions to be facilitated despite being bad in context. in the case of FK, these distractor continuations are apparently facilitated because they're in the same semantic category as the expected item. we can test whether LMs, like human brain responses, assign high probabilities to those distractor continuations 

\subsection{\rr{}: event knowledge and semantic role sensitivity}

Our second set targets event knowledge and semantic role interpretation, and tests sensitivity to impact of role reversals. Table~\ref{tab:chowsents} shows an example item pair from this set. These items come from a human experiment by \newcite{chow2016bag}, which tested the brain's sensitivity to role reversals. 

\begin{table}[h]
\begin{center}
\begin{tabularx}{.46\textwidth}{X | l }
\toprule
Context & Compl.  \\
\midrule
\emph{the restaurant owner forgot which customer the waitress had \_\_\_\_} & \emph{served} \\
\emph{the restaurant owner forgot which waitress the customer had \_\_\_\_} & \emph{served}\\
\bottomrule
\end{tabularx}
\caption{\label{tab:chowsents} Example items from \rr{} dataset (Compl = Context Completion)}
\end{center}
\vspace{-4mm}
\end{table}

\paragraph{Information needed for prediction} Accurate prediction on this set requires a model to interpret semantic roles from sentence syntax, and apply event knowledge about typical interactions between types of entities in the given roles. The set has reversals for each noun pair (shown in Table~\ref{tab:chowsents}) so models must distinguish roles for each order.

%much shorter context now, where in order to make intelligent predictions, the model is expected to interpret semantic roles from syntax, and apply world knowledge about typical interactions between entities. here, the accuracy numbers are informative, but because the cloze values are lower, the comparisons from the N400 experiment are more directly relevant, allowing us to test whether the model can distinguish how good the continuation is based on which roles the nouns are filling

\paragraph{Sensitivity test} The \newcite{chow2016bag} study found that although each completion (e.g., \emph{served}) is good for only one of the noun orders and not the reverse, the N400 shows a similar level of expectation for the target completions regardless of noun order. Our sensitivity test targets this distinction, testing whether LMs will show similar difficulty distinguishing appropriate continuations based on word order and semantic role. Human cloze probabilities show strong sensitivity to the role reversal, with average cloze difference of .233 between good and bad contexts for a given completion.
%We see in Table~\ref{tab:chowsents} that by reversing the positions of \emph{waitress} and \emph{customer}, we reverse those entities' roles relative to the upcoming verb, changing the appropriate completions---\emph{served} is a good completion for the first sentence context, but when the nouns switch position, \emph{served} is no longer sensible. 

\paragraph{Data} The authors provide 120 sentences (60 pairs)---which we filter to 88 final items, removing pairs for which the best completion of either context is not in BERT's single-word vocabulary.

\subsection{\nkf{}: negation}

Our third set targets understanding of the meaning of negation, along with knowledge of category membership. Table~\ref{tab:fssents} shows examples of these test items, which involve absence or presence of negation in simple sentences, with two different completions that vary in truth depending on the negation. These test items come from a human study by \newcite{fischler1983brain}, which examined how human expectations change with the addition of negation. 

\begin{table}[h]
\begin{center}
\begin{tabularx}{.45\textwidth}{X l l }
\toprule
Context & Match  & Mismatch\\
\midrule
\emph{A robin is a \_\_\_\_} & \emph{bird}  & \emph{tree} \\
\emph{A robin is not a \_\_\_\_} & \emph{bird} & \emph{tree} \\
\bottomrule
\end{tabularx}
\caption{\label{tab:fssents} Example items from \fs{} dataset}
\end{center}
\vspace{-4mm}
\end{table}

%now tiny context. accuracy is only relevant for affirmative contexts, in which we see the models are actually very good at associating these nouns with their categories. but the key question here is whether the model can prefer true continuations both with and without negation ... and it definitely can't. when we create contexts more like those the model could have seen/memorized, we now see it working better, but this does not suggest a general handling of the meaning of negation, at very least

\paragraph{Information needed for prediction} Because the negative contexts in these items are highly unconstraining (\emph{A robin is not a \_\_\_\_} ?), prediction accuracy is not a useful measure for the negative contexts. We test prediction accuracy for affirmative contexts only, which allows us to test models' use of hypernym information (\emph{robin} = \emph{bird}). Targeting of negation happens in the sensitivity test.

\paragraph{Sensitivity test} The~\newcite{fischler1983brain} study found that although the N400 shows more expectation for \emph{true} completions in affirmative sentences (e.g., \emph{A robin is a \underline{bird}}), it fails to adjust to negation, showing more expectation for \emph{false} continuations in negative sentences (e.g., \emph{A robin is not a \underline{bird}}). Our sensitivity test targets this distinction, testing whether LMs will show similar insensitivity to impacts of negation. Note that here we use truth judgments rather than cloze probability as an indication of the quality of a completion.
%The addition of \emph{not} in the second sentence reverses the set of completions that constitute acceptable (true) sentences. For this reason, we expect that humans would likely complete the first sentence with \emph{bird}, but never with \emph{tree}. In the case of the second sentence, \emph{tree} becomes the true continuation, while \emph{bird} is false. In this case, tempting inappropriate completions exist primarily for the negative sentences: \emph{A robin is not a bird} is a false sentence, but \emph{robin} and \emph{bird} are likely to have a high affinity due to their category match. 

\paragraph{Data} Fischler et al.~provide the list of 18 subject nouns and 9 category nouns that they use for their sentences, which we use to generate a comparable dataset, for a total of 72 items.\footnote{The one modification that we make to the original subject noun list is a substitution of the word \emph{salmon} for \emph{bass} within the category of fish---because \emph{bass} created lexical ambiguity that was not interesting for our purposes here.} We refer to these 72 simple sentences as \fs{}. All target words are in BERT's single-word vocabulary.

\paragraph{Supplementary items} In a subsequent study, \newcite{nieuwland2008truth} followed up on the~\newcite{fischler1983brain} experiment, creating affirmative and negative sentences chosen to be more ``natural ... for somebody to say'', and contrasting these with affirmative and negative sentences chosen to be less natural. ``Natural'' items include examples like \emph{Most smokers find that quitting is (not) very (difficult/easy)}, while items designed to be less natural include examples like \emph{Vitamins and proteins are (not) very (good/bad)}. The authors share 16 base contexts, corresponding to 64 additional items, which we add to the 72 above for additional comparison. All target words are in BERT's single-word vocabulary. We refer to these supplementary 64 items, designed to test effects of naturalness, as \nk{}.

\section{Experiments}

As a case study, we use these three diagnostics to examine the predictive capacities of the pre-trained BERT model~\cite{devlin2018bert}, which has been the basis of impressive performance across a wide range of tasks. BERT is a deep bidirectional transformer network \cite{vaswani2017attention} pre-trained on tasks of masked language modeling (predicting masked words given bidirectional context) and next-sentence prediction (binary classification of whether two sentences are a sequence). We test two versions of the pre-trained model: \BB{} and \BL{} (uncased). These versions have the same basic architecture, but \BL{} has more parameters---in total, \BB{} has 110M parameters, and \BL{} has 340M. We use the PyTorch BERT implementation with masked language modeling parameters for generating word predictions.\footnote{\url{https://github.com/huggingface/pytorch-pretrained-BERT}}

For testing, we process our sentence contexts to have a \texttt{$[$MASK$]$} token---also used during BERT's pre-training---in the target position of interest. We then measure BERT's predictions for this \texttt{$[$MASK$]$} token's position. Following~\newcite{goldberg2019assessing}, we also add a \texttt{$[$CLS$]$} token to the start of each sentence to mimic BERT's training conditions. 

%We then examine how the context influences BERT's predictions about the target position. This set-up is designed to imitate experiments that have been run on humans, with these same materials, to test expectations formed about upcoming words. 

BERT differs from traditional left-to-right language models, and from real-time human predictions, in being a bidirectional model able to use information from both left and right context. This difference should be neutralized by the fact that our items provide all information in the left context---however, in our experiments here, we do allow one advantage for BERT's bidirectionality: we include a period and a \texttt{$[$SEP$]$} token after each \texttt{$[$MASK$]$} token, to indicate that the target position is followed by the end of the sentence. We do this in order to give BERT the best possible chance of success, by maximizing the chance of predicting a single word rather than the start of a phrase. Items for these experiments thus appear as follows:\\
%\td{Decide about using SEP and period} 

\emph{$[$CLS$]$ The restaurant owner forgot which customer the waitress had $[$MASK$]$ . $[$SEP$]$} \\

Logits produced by the language model for the target position are softmax-transformed to obtain probabilities comparable to human cloze probability values for those target positions.\footnote{Human cloze probabilities are importantly different from true probabilities over a vocabulary, making these values not directly comparable. However, cloze provides important indication---the best indication we have---of how much a context constrains human expectations toward a continuation, so we do at times loosely compare these two types of values.} 

\begin{table*}[t]
\begin{center}
\begin{tabularx}{.7\textwidth}{ X  l l l X}
\toprule
  & Orig & Shuf & Trunc & Shuf + Trunc \\
 \midrule
BERT\textsubscript{BASE} $k=1$    & 23.5 & 14.1 $\pm$ 3.1 & 14.7 & 8.1 $\pm$ 3.4\\
BERT\textsubscript{LARGE} $k=1$ & 35.3 & 17.4 $\pm$ 3.5 & 17.6 & 10.0 $\pm$ 3.0\\
BERT\textsubscript{BASE} $k=5$    & 52.9 & 36.1 $\pm$ 2.8 & 35.3 & 22.1 $\pm$ 3.2\\
BERT\textsubscript{LARGE} $k=5$ & 52.9 & 39.2 $\pm$ 3.9 & 32.4 & 21.3 $\pm$ 3.7\\
\bottomrule
\end{tabularx}
\caption{\label{tab:fkacc} \fk{} word prediction accuracies (with and without sentence perturbations). `Shuf' = first sentence shuffled, `Trunc' = second sentence truncated to two words before target.}
\end{center}
\vspace{-3mm}
\end{table*}

\section{Results for \fk{}}\label{sec:fk}

First we report BERT's results on the \fk{} test targeting common sense, pragmatic reasoning, and sensitivity within semantic category.

\subsection{Word prediction accuracies}

We define accuracy as percentage of items for which the ``expected'' completion is among the model's top $k$ predictions, with $k=1$ and $k=5$. 

Table~\ref{tab:fkacc} (``Orig'') shows accuracies of \BB{} and \BL{}. For accuracy at $k=1$, \BL{} soundly outperforms \BB{} with correct predictions on just over a third of contexts. Expanding to $k=5$, the models converge on the same accuracy, identifying the expected completion for about half of contexts.\footnote{Note that word accuracies are computed by context, so these accuracies are out of the 34 base contexts.}

Because common sense and pragmatic reasoning are non-trivial concepts to pin down, it is worth asking to what extent BERT can achieve this performance based on simpler cues like word identities or n-gram context. To test importance of word order, we shuffle the words in each item's first sentence, garbling the message but leaving all individual words intact (``Shuf'' in Table~\ref{tab:fkacc}). To test adequacy of n-gram context, we truncate the second sentence, removing all but the two words preceding the target word (``Trunc'')---leaving generally enough syntactic context to identify the part of speech, as well as some sense of semantic category (on top of the thematic setup of the first sentence), but removing other information from that second sentence. We also test with both perturbations together (``Shuf + Trunc''). Because different shuffled word orders give rise to different results, for the ``Shuf'' and ``Shuf + Trunc'' settings we show mean and standard deviation from 100 runs. 

Table~\ref{tab:fkacc} shows the accuracies as a result of these perturbations. One thing that is immediately clear is that the BERT model is indeed making use of information provided by the word order of the first sentence, and by the more distant content of the second sentence, as each of these individual perturbations causes a notable drop in accuracy. It is worth noting, however, that with each perturbation there is a subset of items for which BERT's accuracy remains intact. Unsurprisingly, many of these items are those containing particularly distinctive words associated with the target, such as \emph{checkmate} (\emph{chess}), \emph{touchdown} (\emph{football}), and \emph{stone-washed} (\emph{jeans}). This suggests that some of BERT's success on these items may be attributable to simpler lexical or n-gram information. In Section~\ref{sec:fkqual} we take a closer look at some more difficult items that seemingly avoid such loopholes.

\subsection{Completion sensitivity}

Next we test BERT's ability to prefer expected completions over inappropriate completions of the same semantic category. We first test this by simply measuring the percentage of items for which BERT assigns a higher probability to the good completion (e.g., \emph{lipstick} from Table~\ref{tab:fksents}) than to either of the inappropriate completions (e.g., \emph{mascara, bracelet}). Table~\ref{tab:fkgoodbad} shows the results. We see that \BB{} assigns the highest probability to the expected completion in 73.5\% of items, while \BL{} does so for 79.4\%---a solid majority, but with a clear portion of items for which an inappropriate, semantically-related target does receive a higher probability than the appropriate word.

\begin{table}[h]
\begin{center}
\begin{tabularx}{.45\textwidth}{ l l l}
\toprule
 & Prefer good & w/ .01 thresh \\
\midrule
BERT\textsubscript{BASE}  & 73.5 & 44.1  \\
BERT\textsubscript{LARGE}  & 79.4 & 58.8  \\
\bottomrule
\end{tabularx}
\caption{\label{tab:fkgoodbad} Percent of \fk{} items with good completion assigned higher probability than bad}
\end{center}
\vspace{-4mm}
\end{table}

 \begin{table*}[t]
\begin{center}
\begin{tabularx}{.95\textwidth}{X | l }
\toprule
Context & \BL{} predictions  \\
\midrule
\emph{Pablo wanted to cut the lumber he had bought to make some shelves. He asked his neighbor if he could borrow her \_\_\_\_} & \emph{car, house, room, truck, apartment} \\
\emph{The snow had piled up on the drive so high that they couldn't get the car out. When Albert woke up, his father handed him a \_\_\_\_ } & \emph{note, letter, gun, blanket, newspaper} \\
\emph{At the zoo, my sister asked if they painted the black and white stripes on the animal. I explained to her that they were natural features of a \_\_\_\_} & \emph{cat, person, human, bird, species} \\
\bottomrule
\end{tabularx}
\caption{\label{tab:fkerr} \BL{} top word predictions for selected \fk{} items }
\end{center}
\vspace{-4mm}
\end{table*}

We can make our criterion slightly more stringent if we introduce a threshold on the probability difference. The average cloze difference between good and bad completions is about .74 for the data from which these items originate, reflecting a very strong human sensitivity to the difference in completion quality. To test the proportion of items in which BERT assigns more substantially different probabilities, we filter to items for which the good completion probability is higher by greater than .01---a threshold chosen to be very generous given the significant average cloze difference. With this threshold, the sensitivity drops noticeably---\BB{} shows sensitivity in only 44.1\% of items, and \BL{} shows sensitivity in only 58.8\%. These results tell us that although the models are able to prefer good completions to same-category bad completions in a majority of these items, the difference is in many cases very small, suggesting that this sensitivity falls short of what we see in human cloze responses.

\subsection{Qualitative examination of predictions}\label{sec:fkqual}

We see above that the BERT models are able to identify the correct word completions in approximately half of \fk{} items, and that the models are able to prefer good completions to semantically-related inappropriate completions in a majority of items, though with notably weaker sensitivity than humans. To better understand the models' weaknesses, in this section we examine predictions made when the models fail.

Table~\ref{tab:fkerr}  shows three example items along with the top five predictions of \BL{}. In each case, BERT provides completions that are sensible in the context of the second sentence, but that fail to take into account the context provided by the first sentence---in particular, the predictions show no evidence of having been able to infer the relevant information about the situation or object described in the first sentence. For instance, we see in the first example that BERT has correctly zeroed in on things that one might borrow, but it fails to infer that the thing to be borrowed is something to be used for cutting lumber. Similarly, BERT's failure to detect the snow-shoveling theme of the second item makes for an amusing set of non sequitur completions. Finally, the third example shows that BERT has identified an animal theme (unsurprising, given the words \emph{zoo} and \emph{animal}), but it is not applying the phrase \emph{black and white stripes} to identify the appropriate completion of \emph{zebra}. Altogether, these examples illustrate that with respect to the target capacities of commonsense inference and pragmatic reasoning, BERT fails in these more challenging cases.

\begin{table*}[t]
\begin{center}
\begin{tabularx}{\textwidth}{X | p{4.5cm} | p{4.5cm} }
\toprule
Context & \BB{} predictions & \BL{} predictions  \\
\midrule
\emph{the camper reported which girl the bear had \_\_\_\_} & \emph{taken, killed, attacked, bitten, picked}   & \emph{attacked, killed, eaten, taken, targeted}\\
\emph{the camper reported which bear the girl had \_\_\_\_} & \emph{taken, killed, fallen, bitten, jumped} & \emph{taken, left, entered, found, chosen}  \\
\midrule
\emph{the restaurant owner forgot which customer the waitress had \_\_\_\_} & \emph{served, hired, brought, been, taken} &  \emph{served, been, delivered, mentioned, brought} \\
\emph{the restaurant owner forgot which waitress the customer had \_\_\_\_} & \emph{served, been, chosen, ordered, hired} & \emph{served, chosen, called, ordered, been} \\
\bottomrule
\end{tabularx}
\caption{\label{tab:chowerr} \BB{} and \BL{} top word predictions for selected \rr{} sentences}
\end{center}
\end{table*}

\section{Results for \rr{}}\label{sec:rr}

Next we turn to the \rr{} test of semantic role sensitivity and event knowledge.

\subsection{Word prediction accuracies}\label{sec:rracc}

We again define accuracy by presence of a top cloze item within the model's top $k$ predictions. Table~\ref{tab:chowacc} (``Orig'') shows the accuracies for \BL{} and \BB{}. For $k=1$, accuracies are very low, with \BB{} slightly outperforming \BL{}. When we expand to $k=5$, accuracies predictably increase, and \BL{} now outperforms \BB{} by a healthy margin.

\begin{table}[h]
\begin{center}
\begin{tabularx}{.475\textwidth}{ l l l l l}
\toprule
 & Orig & -Obj & -Sub & -Both\\
 \midrule
BERT\textsubscript{BASE} $k$=$1$ & 14.8 & 12.5 & 12.5 & 9.1 \\
BERT\textsubscript{LARGE} $k$=$1$ & 13.6 & 5.7  & 6.8 & 4.5 \\
BERT\textsubscript{BASE} $k$=$5$ & 27.3 & 26.1 & 22.7 & 18.2 \\
BERT\textsubscript{LARGE} $k$=$5$ & 37.5 & 18.2 & 21.6 & 14.8 \\
\bottomrule
\end{tabularx}
\caption{\label{tab:chowacc} \rr{} word prediction accuracies (with and without sentence perturbations). `-Obj' = generic object, `-Subj' = generic subject, `-Both' = generic object and subject.}
\end{center}
\vspace{-2mm}
\end{table}

%\toprule
%   & BERT\textsubscript{LARGE}  predictions  \\
%\midrule
%\emph{the camper reported which girl the bear had  \_\_\_\_} & \emph{attacked, killed, eaten, taken, targeted} \\
%\emph{the camper reported which bear the girl had \_\_\_\_} & \emph{taken, left, entered, found, chosen} \\
%\midrule
%\emph{the restaurant owner forgot which customer the waitress had \_\_\_\_} & \emph{served, been, delivered, mentioned, brought} \\
%\emph{the restaurant owner forgot which waitress the customer had  \_\_\_\_} & \emph{served, chosen, called, ordered, been} \\
%\bottomrule
%\end{tabularx}
%\caption{\label{tab:chowerrlg} \BB{} and \BL{} predictions on selected \rr{} sentences}
%\end{center}
%\end{table*}

To test the extent to which BERT is relying on the individual nouns in the context, we try two different perturbations of the contexts: removing the information from the object 
(\emph{which \underline{customer} the waitress ...}), and removing the information from the subject (\emph{which customer the \underline{waitress}...}), 
in each case by replacing the noun with a generic substitute. We choose \emph{one} and \emph{other} as substitutions for the object and subject, respectively.

Table~\ref{tab:chowacc} shows the results with each of these perturbations individually and together. We observe several notable patterns. First, removing either the object (``-Obj'') or the subject (``-Sub'') has relatively little effect on the accuracy of \BB{} for either $k=1$ or $k=5$. This is quite different from what we see with \BL{}, the accuracy of which drops substantially when the object or subject information is removed. These patterns suggest that \BB{} is less dependent upon the full detail of the subject-object structure, instead relying primarily upon one or the other of the participating nouns for its verb predictions. \BL{}, on the other hand, appears to make heavier use of both nouns, such that loss of either one causes non-trivial disruption in the predictive accuracy.

\begin{table}[t]
\begin{center}
\begin{tabularx}{.475\textwidth}{ l l l l l}
\toprule
                   & $\leq$.17 & $\leq$.23  & $\leq$.33 & $\leq$.77 \\
\midrule
BERT\textsubscript{BASE}  $k$=$1$ & 12.0 & 17.4 & 17.4 & 11.8 \\
BERT\textsubscript{LARGE} $k$=$1$ & 8.0 & 4.3 & 17.4 & 29.4 \\
BERT\textsubscript{BASE} $k$=$5$ & 24.0 & 26.1 & 21.7 & 41.1 \\
BERT\textsubscript{LARGE} $k$=$5$ & 28.0 & 34.8 & 39.1 & 52.9 \\
\bottomrule
\end{tabularx}
\caption{\label{tab:clozebins} Accuracy of predictions in unperturbed \rr{} sentences, binned by max cloze of context}
\end{center}
\vspace{-5mm}
\end{table}

It should be noted that the items in this set are overall less constraining than those in Section~\ref{sec:fk}---humans converge less clearly on the same predictions, resulting in lower average cloze values for the best completions. To investigate the effect of constraint level, we divide items into four bins by top cloze value per sentence. Table~\ref{tab:clozebins} shows the results. With the exception of \BB{} at $k=1$, for which accuracy in all bins is fairly low, it is clear that the highest cloze bin yields much higher model accuracies than the other three bins, suggesting some alignment between how constraining contexts are for humans and how constraining they are for BERT. However, even in the highest cloze bin, when at least a third of humans converge on the same completion, even \BL{} at $k=5$ is only correct in half of cases, suggesting substantial room for improvement.\footnote{This analysis is made possible by the~\newcite{chow2016bag} authors' generous provision of the cloze data for these items, not originally made public with the items themselves.} 

\subsection{Completion sensitivity}

Next we test BERT's sensitivity to role reversals by comparing model probabilities for a given completion (e.g., \emph{served}) in the appropriate versus role-reversed contexts. We again start by testing the percentage of items for which BERT assigns a higher probability to the appropriate than to the inappropriate completion.  As we see in Table~\ref{tab:goodbad}, \BB{} prefers the good continuation in 75\% of items, while \BL{} does so for 86.4\%---comparable to the proportions for \fk{}. However, when we apply our threshold of .01---still generous given the average cloze difference of .233---sensitivity drops more dramatically than on \fk{}, to 31.8\% and 43.2\%. 

\begin{table}[t]
\begin{center}
\begin{tabular}{ l l l}
\toprule
 & Prefer good & w/ .01 thresh \\
\midrule
BERT\textsubscript{BASE}  & 75.0 & 31.8  \\
BERT\textsubscript{LARGE}  & 86.4 & 43.2  \\
\bottomrule
\end{tabular}
\caption{\label{tab:goodbad} Percent of \rr{} items with good completion assigned higher probability than role reversal}
\end{center}
\vspace{-5mm}
\end{table}

%\td{consider removing} To explore these probability differences in greater detail, Figures~\ref{fig:basescat} and~\ref{fig:largescat} show the differences in model probabilities for good versus bad completions, plotted against the differences in the human cloze values for those same pairs. Although the models' probabilities are generally above zero (indicating preference for good completions), we see that most values are clustered near zero. The figures also indicate that the greatest model probability differences occur for items on which humans cloze values are only moderately different, while for the items for which human cloze values diverge most starkly, the probability differences are low.

Overall, these results suggest that BERT is, in a majority of cases of this kind, able to use noun position to prefer good verb completions to bad---however, it is again less sensitive than humans to these distinctions, and it fails to match human word predictions on a solid majority of cases. The model's ability to choose good completions over role reversals (albeit with weak sensitivity) suggests that the failures on word prediction accuracy are not due to inability to distinguish word orders, but rather to a weakness in event knowledge or understanding of semantic role implications.

%\begin{figure*}[t]
%\centering
%\begin{minipage}{0.5\textwidth}
%\centering
%\includegraphics[width =1\textwidth]{prcl-bert-base-uncased}
%\caption{\BB{} }\label{fig:basescat}
%\end{minipage}\hfill
%\begin{minipage}{0.5\textwidth}
%\centering
%\includegraphics[width =1\textwidth]{prcl-bert-large-uncased}
%\caption{\BL{}}\label{fig:largescat}
%\end{minipage}
%\end{figure*}

\subsection{Qualitative examination of predictions}

Table~\ref{tab:chowerr} shows predictions of \BB{} and \BL{} for some illustrative examples. For the \emph{girl}/\emph{bear} items, we see that \BB{} favors continuations like \emph{killed} and \emph{bitten} with \emph{bear} as subject, but also includes these continuations with \emph{girl} as subject. \BL{}, by contrast, excludes these continuations when \emph{girl} is the subject. 

In the second pair of sentences we see that the models choose \emph{served} as the top continuation under both word orders, even though for the second word order this produces an unlikely scenario. In both cases, the model's assigned probability for \emph{served} is much higher for the appropriate word order than the inappropriate one---a difference of .6 for \BL{} and .37 for \BB{}---but it is noteworthy that no more semantically appropriate top continuation is identified by either model for \emph{which waitress the customer had \_\_\_\_}. 

As a final note, although the continuations are generally impressively grammatical, we see exceptions in the second \emph{bear}/\emph{girl} sentence---both models produce completions of questionable grammaticality (or at least questionable use of selection restrictions), with sentences like \emph{which bear the girl had fallen} from \BB{}, and \emph{which bear the girl had entered} from \BL{}. 

\section{Results for \nkf{}}\label{sec:neg}

Finally, we turn to the \nkf{} test of negation and category membership.

\subsection{Word prediction accuracies}
 
We start by testing the ability of BERT to predict correct category continuations for the affirmative contexts in \fs{}. Table~\ref{tab:fsacc} shows the accuracy results for these affirmative sentences.

\begin{table}[h]
\begin{center}
\begin{tabular}{ l c}
\toprule
 & Accuracy \\
\midrule
BERT\textsubscript{BASE} $k=1$ & 38.9  \\
BERT\textsubscript{LARGE}  $k=1$ & 44.4 \\
BERT\textsubscript{BASE}  $k=5$ & 100  \\
BERT\textsubscript{LARGE}  $k=5$ & 100  \\
\hline
\end{tabular}
\caption{\label{tab:fsacc} Accuracy of word predictions in \fs{} affirmative sentences}
\end{center}
\vspace{-5mm}
\end{table}

We see that for $k=5$, the correct category is predicted for 100\% of affirmative items, suggesting an impressive ability of both BERT models to associate nouns with their correct immediate hypernyms. We also see that the accuracy drops substantially when assessed on $k=1$. Examination of predictions reveals that these errors consist exclusively of cases in which BERT completes the sentence with a repetition of the subject noun, e.g., \emph{A daisy is a daisy}---which is certainly true, but which is not a likely or informative sentence. 

\subsection{Completion sensitivity}

We next assess BERT's sensitivity to the meaning of negation, by measuring the proportion of items in which the model assigns higher probabilities to true completions than to false ones.

\begin{table}[h]
\begin{center}
\begin{tabularx}{.4\textwidth}{ l l l }
\toprule
 & Affirmative &  Negative  \\
\midrule
BERT\textsubscript{BASE} & 100 &  0.0   \\
BERT\textsubscript{LARGE} & 100 &  0.0  \\
\bottomrule
\end{tabularx}
\caption{\label{tab:fsgoodbad} Percent of \fs{} items with true completion assigned higher probability than false}
\end{center}
\vspace{-4mm}
\end{table}

Table~\ref{tab:fsgoodbad} shows the results, and the pattern is stark. When the statement is affirmative (\emph{A robin is a \_\_\_\_}), the models assign higher probability to the true completion in 100\% of items. Even with the threshold of .01---which eliminated many comparisons on \fk{} and \rr{}---all items pass but one (for \BB{}), suggesting a robust preference for the true completions.

However, in the negative statements (\emph{A robin is not a \_\_\_\_}), BERT prefers the true completion in 0\% of items, assigning the higher probability to the false completion in every case. This shows a strong insensitivity to the meaning of negation, with BERT  preferring the category match completion every time, despite its falsity. 

\begin{table*}[t]
\begin{center}
\begin{tabularx}{.8\textwidth}{l | X }
\toprule
Context & \BL{} predictions  \\
\midrule
\emph{A robin is a \_\_\_\_ } & \emph{bird, robin, person, hunter, pigeon} \\
\emph{A daisy is a \_\_\_\_}   & \emph{daisy, rose, flower, berry, tree} \\
\emph{A hammer is a \_\_\_\_}   & \emph{hammer, tool, weapon, nail, device} \\
\emph{A hammer is an \_\_\_\_}   & \emph{object, instrument, axe, implement, explosive} \\
\midrule
\emph{A robin is not a \_\_\_\_}   & \emph{robin, bird, penguin, man, fly} \\
\emph{A daisy is not a \_\_\_\_}   & \emph{daisy, rose, flower, lily, cherry}\\
\emph{A hammer is not a \_\_\_\_}   & \emph{hammer, weapon, tool, gun, rock} \\
\emph{A hammer is not an \_\_\_\_ }  & \emph{object, instrument, axe, animal, artifact} \\
\bottomrule
\end{tabularx}
\caption{\label{tab:fspreds} \BL{} top word predictions for selected \fs{} sentences }
\end{center}
\vspace{-4mm}
\end{table*}

\subsection{Qualitative examination of predictions}

Table~\ref{tab:fspreds} shows examples of the predictions made by \BL{} in positive and negative contexts. We see a clear illustration of the phenomenon suggested by the results above: for affirmative sentences, BERT produces generally true completions (at least in the top two)---but these completions remain largely unchanged after negation is added, resulting in many blatantly untrue completions. 

Another interesting phenomenon that we can observe in Table~\ref{tab:fspreds} is BERT's sensitivity to the nature of the determiner (\emph{a} or \emph{an}) preceding the masked word. This determiner varies depending on whether the upcoming target begins with a vowel or a consonant (for instance, our mismatched category paired with \emph{hammer} is \emph{insect}) and so the model can potentially use this cue to filter the predictions to those starting with either vowels or consonants. How effectively does BERT use this cue? The predictions indicate that BERT is for the most part extremely good at using this cue to limit to words that begin with the right type of letter. There are certain exceptions (e.g., \emph{An ant is not a ant}), but these are in the minority.

\subsection{Increasing naturalness}

The supplementary \nk{} items allow us to examine further the model's handling of negation, with items designed to test the effect of ``naturalness''.  When we present BERT with this new set of sentences, the model does show an apparent change in sensitivity to the negation. \BB{} assigns true statements higher probability than false for 75\% of natural sentences (``NT''), and \BL{} does so for 87.5\% of natural sentences. By contrast, the models each show preference for true statements in only 37.5\% of items designed to be less natural (``LN''). Table~\ref{tab:nkgoodbad} shows these sensitivities broken down by affirmative and negative conditions. Here we see that in the natural sentences, BERT prefers true statements for both affirmative and negative contexts---by contrast, the less natural sentences show the pattern exhibited on \fs{}, in which BERT prefers true statements in a high proportion of affirmative sentences, and in 0\% of negative sentences, suggesting that once again BERT is defaulting to category matches with the subject. 
\begin{table}[h]
\begin{center}
\begin{tabularx}{.48\textwidth}{ l X X X X}
\toprule
 & Aff. NT & Neg. NT & Aff. LN & Neg. LN \\
\midrule
\BB{} & 62.5 & 87.5  & 75.0 & 0.0  \\
\BL{} & 75.0 & 100  & 75.0 & 0.0 \\
\bottomrule
\end{tabularx}
\caption{\label{tab:nkgoodbad} Percent of \nk{} with true continuation given higher probability than false. Aff = affirmative, Neg = negative. NT = natural, LN = less natural.}
\end{center}
\vspace{-4mm}
\end{table}

Table~\ref{tab:nkpreds} contains \BL{} predictions on two pairs of sentences from the ``Natural'' sentence set. It is worth noting that even when BERT's first prediction is appropriate in the context, the top candidates often contradict each other (e.g., \emph{difficult} and \emph{easy}). We also see that even with these natural items, sometimes the negation is not enough to reverse the top completions, as with the second pair of sentences, in which the fast food dinner both is and isn't a romantic first date.

\begin{table*}[t]
\begin{center}
\begin{tabularx}{\textwidth}{l | X }
\toprule
Context & \BL{} predictions  \\
\midrule
\emph{Most smokers find that quitting is very \_\_\_\_ } & \emph{difficult, easy, effective, dangerous, hard} \\
\emph{Most smokers find that quitting isn't very \_\_\_\_} & \emph{effective, easy, attractive, difficult, succcessful} \\
\emph{A fast food dinner on a first date is very \_\_\_\_} & \emph{good, nice, common, romantic, attractive} \\
\emph{A fast food dinner on a first date isn't very \_\_\_\_} & \emph{nice, good, romantic, appealing, exciting} \\
\bottomrule
\end{tabularx} 
\caption{\label{tab:nkpreds} \BL{} top word predictions for selected \nk{} sentences }
\end{center}
\vspace{-4mm}
\end{table*}

%Fischler http://www.coli.uni-saarland.de/courses/nondensity15/fischler83.pdf
%When the truth is not too hard to handle https://projects.iq.harvard.edu/files/kuperberglab/files/nieuwland_kuperberg_negationexamples.pdf
%https://projects.iq.harvard.edu/files/kuperberglab/files/nieuwlandkuperberg_psychsci_08.pdf

\section{Discussion}

Our three diagnostics allow for a clarified picture of the types of information used for predictions by pre-trained BERT models. On \fk{}, we see that both models can predict the best completion approximately half the time (at $k=5$), and that both models rely non-trivially on word order and full sentence context. However, successful predictions in the face of perturbations also suggest that some of BERT's success on these items may exploit loopholes, and when we examine predictions on challenging items, we see clear weaknesses in the commonsense and pragmatic inferences targeted by this set. Sensitivity tests show that BERT can also prefer good completions to bad semantically-related completions in a majority of items, but many of these probability differences are very small, suggesting that the model's sensitivity is much less than that of humans.

% Testing BERT on Set 1, we see that both models are able to use the contextual cues to predict the best completion nearly half the time (at $k=5$), a fairly impressive feat given how difficult one might expect these sentences to be. Perturbations reveal that both models are relying non-trivially on word order and full sentence context, with \BL{} at $k=1$ particularly impacted by these changes. Examining completion probabilities, we see that BERT's probability assignments typically favor good over bad completions, though with much less sensitivity than humans. When we examine BERT's predictions more closely, we see particular weaknesses emerge, particularly in inferring the relevance of particular items when that relevance is only implied. What we see, then, is that BERT can in many cases make impressively correct predictions, and it is on average able to sort good continuations from bad ones among those presented to it in this set, but there are many aspects of situation description clues that are able to be used by humans, but that BERT appears to miss entirely. 

On \rr{}, BERT's accuracy in matching top human predictions is much lower, with \BL{} at only 37.5\% accuracy. 
%This drop in accuracy is in part attributable to lower agreement among humans about best completions for these items, but the drop in accuracy may also be due to the subtler nature of the distinctions in this test set, which holds word content constant and simply reverses the positions of noun phrases. 
Perturbations reveal interesting model differences, suggesting that \BL{} has more sensitivity than \BB{} to the interaction between subject and object nouns. Sensitivity tests show that both models are typically able to use noun position to prefer good completions to role reversals, but the differences are on average even smaller than on \fk{}, indicating again that model sensitivity to the distinctions is less than that of humans. The models' general ability to distinguish role reversals suggests that the low word prediction accuracies are not due to insensitivity to word order \emph{per se}, but rather to weaknesses in event knowledge or understanding of semantic role implications.

Finally, \nkf{} allows us to zero in with particular clarity on a divergence between BERT's predictive behavior and what we might expect from a model using all available information about word meaning and truth/falsity. When presented with simple sentences describing category membership, BERT shows a complete inability to prefer true over false completions for negative sentences. The model shows an impressive ability to associate subject nouns with their hypernyms, but when negation reverses the truth of those hypernyms, BERT continues to predict them nonetheless. By contrast, when presented with sentences that are more ``natural'', BERT does reliably prefer true completions to false, with or without negation. Although these latter sentences are designed to differ in naturalness, in all likelihood it is not naturalness \emph{per se} that drives the model's relative success on them---but rather a higher frequency of these types of statements in the training data.

The latter result in particular serves to highlight a stark, but ultimately unsurprising, observation about what these pre-trained language models bring to the table. While the function of language processing for humans is to compute meaning and make judgments of truth, language models are trained as predictive models---they will simply leverage the most reliable cues in order to optimize their predictive capacity. For a phenomenon like negation, which is often not conducive to clear predictions, such models may not be equipped to learn the implications of this word's meaning.

%
%A couple of notes:
%prediction of words is not necessarily something we should expect humans to be state of the art at -- we expect them to be better at actual understanding, but predicting words is a different thing
%
%to the extent that understanding allows for better predictions --- something that we expect to be the case in these sentences --- we do have a sense that humans should be better at picking up on the relevant cues and using them to predict words based on understanding of the situation
%
%
%\begin{enumerate}
%\item BERT generally extremely good from a syntactic perspective --- always correct POS as far as I can tell, and seems also to handle a/an alternation most of the time
%\item BERT does seem in general to give more sensible completions for things that likely occur more in the training data---things that seem like they might be actual phrases that could have occurred approximately in the training data
%\item It really likes ``killed'' as a completion for the WY stimuli ...
%\item It's definitely happy to produce blatantly untrue statements.
%\item That said, it brings in the right category for all but one of the Fischler nouns (making it correct for the affirmative statements).
%\item BERT is a cheap date -- thinks a fast food date is romantic.
%\end{enumerate}

\section{Conclusion}

In this paper we have introduced a suite of diagnostic tests for language models, to better our understanding of the linguistic competencies acquired by pre-training via language modeling. We draw our tests from psycholinguistic studies, allowing us to target a range of linguistic capacities by testing word prediction accuracies and sensitivity of model probabilities to linguistic distinctions. As a case study, we apply these tests to analyze strengths and weaknesses of the popular BERT model, finding that it shows sensitivity to role reversal and same-category distinctions, albeit less than humans, and it succeeds with noun hypernyms, but it struggles with challenging inferences and role-based event prediction---and it shows clear failures with the meaning of negation. We make all test sets and experiment code available (see Footnote~\ref{fn:link}), for further experiments.

The capacities targeted by these test sets are by no means comprehensive, and future work can build on the foundation of these datasets to expand to other aspects of language processing. Because these sets are small, we must also be conservative in the strength of our conclusions---different formulations may yield different performance, and future work can expand to verify the generality of these results. 
%\td{As with any dataset, it is important to keep in mind that different settings that test similar phenomena may yield different results.} 
In parallel, we hope that the weaknesses highlighted by these diagnostics can help to identify areas of need for establishing robust and generalizable models for language understanding.

\section*{Acknowledgments}

We would like to thank Tal Linzen, Kevin Gimpel, Yoav Goldberg, Marco Baroni, and several anonymous reviewers for valuable feedback on earlier versions of this paper. We also thank members of the Toyota Technological Institute at Chicago for useful discussion of these and related issues.

\bibliography{tacl2018}
\bibliographystyle{acl_natbib}

\end{document}